\documentclass{article}

\usepackage[dblblindworkshop, final]{neurips_2026}

\usepackage[utf8]{inputenc} 
\usepackage[T1]{fontenc}    
\usepackage{hyperref}       
\usepackage{url}            
\usepackage{booktabs}       
\usepackage{amsfonts}       
\usepackage{nicefrac}       
\usepackage{microtype}      
\usepackage{xcolor}         

\usepackage{amsmath}
\usepackage{graphicx}
\usepackage{amssymb}
\usepackage{subcaption}

\title{When Accuracy Is Not Enough: Evaluating Trustworthiness in High-Stakes Lease Interpretation}
\workshoptitle{TAE (Trust-AI-Eval): Can We Trust AI Evaluation?}

\author{%
  Bin Liang$^{1}$\thanks{Corresponding author: \texttt{bin.liang@uts.edu.au}},
  Yuxi Lu$^{1,2}$,
  Jin Wang$^{1}$,
  Zhidong Li$^{1}$ \\[6pt]
  $^{1}$University of Technology Sydney, Australia \\
  $^{2}$University of New South Wales, Australia
}

\begin{document}

\maketitle

\begin{abstract}
Large language models are increasingly used to explain legal and contractual information, but standard evaluations centred on answer accuracy can miss failures that are consequential in high-stakes settings. In lease interpretation, a response may appear correct while relying on inapplicable legal material, unsupported evidence, or an inappropriate decision to answer. We present an evaluation framework that characterises trustworthiness along three dimensions: legal correctness, evidence integrity, and decision safety, together with explicit hard-failure criteria for serious errors. We study this framework in New South Wales residential tenancy using a 2,000-question benchmark and a stepwise comparison of progressively safeguarded system configurations, from direct generation to applicability-aware retrieval, evidence-first generation, evidence checking, and selective response behaviour. On the protected test set, answer accuracy increases only from 82.3\% to 84.0\%, while the Trustworthy Response Rate (TRR) rises from 58.7\% to 82.3\% and the Hard Failure Rate (HFR) falls from 20.7\% to 5.7\%. Two configurations with identical 83.7\% Answer Accuracy achieve different TRR (72.0\% versus 78.0\%) and HFR (11.7\% versus 8.0\%). These results show that configurations that appear equivalent under answer accuracy can differ substantially under response level trustworthiness measures.
\end{abstract}

\section{Introduction}

Large language models (LLMs) are increasingly used to support legal information seeking, document analysis, and question answering. Recent
benchmarks show that modern LLMs can perform a broad range of legal reasoning tasks \citep{guha2023legalbench}, yet strong task performance does not imply that their outputs are reliable enough for high-stakes use. Legal hallucinations remain common, and models can produce confident responses that misstate legal facts, accept incorrect premises, or fail to recognise their own uncertainty \citep{dahl2024legalfictions}. These risks are particularly important in lease interpretation, where an apparently plausible response may
affect how a renter understands contractual obligations, procedural options, or whether further assistance is needed.

The resulting challenge is partly a problem of \emph{evaluation}. Standard language-model evaluations often summarise performance using aggregate measures such as accuracy or F1. Such measures remain useful, but they capture only part of the behaviour relevant to deployment. Broader evaluation frameworks have therefore argued for measuring multiple dimensions of model behaviour rather than relying on a single performance score \citep{liang2023helm}. More recently, work on construct validity has shown that benchmark conclusions can be misleading when the selected tasks and metrics do not adequately represent the phenomenon that an evaluation claims to measure \citep{bean2025construct}. For high-stakes legal information, this raises a basic question: \emph{what does an accuracy score actually tell us about whether a response is trustworthy enough to show to a user?}

Lease interpretation exposes several failure modes that conventional answer accuracy does not distinguish. A response may reach a broadly correct conclusion while relying on material from the wrong jurisdiction or tenancy context. It may cite a valid source whose content does not support the associated claim. Conversely, a response may be factually plausible but should not have been given because a decisive fact is missing or the request falls outside the supported scope. Prior work on citation evaluation has similarly shown that the presence of citations is distinct from whether those citations adequately support generated claims \citep{gao2023alce}. Work on abstention further demonstrates that knowing when \emph{not} to answer remains a difficult and separately measurable capability, even for strong LLMs \citep{kirichenko2025abstentionbench}.

We study these issues through New South Wales residential tenancy as a high stakes case study. Rather than treating trustworthiness as a single weighted score, we evaluate system behaviour along three complementary dimensions: \textbf{legal correctness}, which measures whether the relevant legal issue and context are handled correctly; \textbf{evidence integrity}, which measures whether supporting material is traceable, applicable, and actually supports the claims made; and \textbf{decision safety}, which measures whether the system appropriately answers, requests clarification, abstains, or refers. We additionally define explicit \emph{hard failures} for
errors that should prevent a response from being regarded as trustworthy, including materially incorrect legal information, unsupported legal claims, wrong context evidence, and unsafe decisions to answer. 

Our evaluation uses a 2,000 question NSW residential tenancy benchmark covering five broad tenancy topics and supporting analysis across question types, risk levels, and response requirements. We compare progressively safeguarded system configurations, beginning with \textbf{Direct Generation} and incrementally adding \textbf{Applicability Aware Retrieval}, \textbf{Evidence First Generation}, \textbf{Evidence Checked Generation}, and the \textbf{Full Safeguarded System}. This stepwise design enables comparison of how successive safeguards are associated with changes in the reported trustworthiness measures. The results show a substantial divergence between conventional accuracy and the broader trustworthiness evaluation. Across the protected test set, Answer Accuracy increases only from 82.3\% to 84.0\%, while Trustworthy Response Rate (TRR) increases from 58.7\% to 82.3\% and Hard Failure Rate (HFR) decreases from 20.7\% to 5.7\%. Evidence Support also rises from 70.2\% to 92.0\%. Topic level analysis further reveals meaningful variation that is obscured
by the aggregate result. 

The contributions of this work are three-fold:
\begin{itemize}
    \item We introduce an operational evaluation protocol for high stakes lease interpretation that separates legal correctness, evidence integrity, and decision safety, together with a non compensatory hard failure gate.

    \item We evaluate the framework on a 2,000 question NSW residential tenancy benchmark spanning five broad tenancy topics and supporting analysis across response requirements and meaningful evaluation slices.

    \item We provide a stepwise safeguard analysis showing substantial gains in trustworthy response rate and evidence support, together with large reductions in serious failures, despite only modest changes in conventional answer accuracy.
\end{itemize}

\section{Related Work}

\subsection{Evaluation of LLMs in high-stakes settings}

Language-model evaluation has increasingly moved beyond single-task accuracy towards broader assessments of robustness, calibration, uncertainty, and deployment-relevant behaviour. HELM, for example, advocates evaluating models across multiple scenarios and metrics rather than reducing performance to a single score \citep{liang2023helm}. Recent work on construct validity further questions whether benchmark tasks and metrics actually measure the capability that an evaluation claims to represent \citep{bean2025construct}. Benchmarks such as TruthfulQA further demonstrate that fluent answer generation does not necessarily imply factual reliability, motivating direct
evaluation of truthfulness alongside conventional task performance \citep{lin2022truthfulqa}. Selective answering introduces an additional dimension: even strong models may fail to recognise underspecified or unanswerable questions, motivating explicit evaluation of abstention behaviour \citep{kirichenko2025abstentionbench}. Our work builds on this perspective but focuses on a high-stakes setting in which correctness, supporting evidence, and the decision to answer can fail independently. Complementary research has explored LLM auditability through verifiable model-update records and compliance-oriented verification mechanisms~\citep{li2025auditablellm}. While such approaches address accountability at the model lifecycle level, our work focuses on trustworthiness at the individual response level.

\subsection{Legal NLP and contract understanding}

Legal NLP benchmarks cover a broad range of tasks, including legal language understanding, contract review, contractual inference, and legal reasoning. LexGLUE provides a standardised collection of legal language-understanding tasks \citep{chalkidis2022lexglue}, while CUAD uses expert annotations to evaluate the identification of legally significant clauses in contracts \citep{hendrycks2021cuad}. ContractNLI extends contract analysis to document-level inference with explicit supporting evidence spans \citep{koreeda2021contractnli}. More recent benchmarks such as LegalBench evaluate diverse forms of legal reasoning \citep{guha2023legalbench}, and CLAUSE stress-tests LLMs using fine-grained discrepancies introduced into contracts \citep{choudhury2026clause}. These resources establish the difficulty of legal language and reasoning, but primarily evaluate task-specific predictions or reasoning outcomes. Our setting instead evaluates the complete response package: the legal conclusion, the applicability and support of its evidence, and whether the system should have
answered in the first place.

\subsection{Evidence support and selective response}

Providing a citation does not by itself establish that a generated claim is adequately supported. ALCE explicitly separates answer quality from citation quality in long-form generation \citep{gao2023alce}, while evidence-based question-answering studies distinguish source quality from whether the answer is attributable to the supplied evidence \citep{schimanski2024evidenceqa}. Evaluation frameworks for retrieval augmented generation similarly separate context relevance and answer quality from whether generated responses are
faithful to the retrieved evidence \citep{es2024ragas,saadfalcon2024ares}. Fine-grained citation evaluation further shows that evidence support may be full, partial, or absent, and that automated support metrics do not behave uniformly across these cases \citep{zhang2024citation}. In parallel, AbstentionBench demonstrates that deciding when not to answer remains a distinct challenge from answering correctly \citep{kirichenko2025abstentionbench}. We distinguish evidence integrity, including traceability, applicability, and claim support, from decision safety, which evaluates whether the appropriate behaviour is to answer, clarify, abstain, or refer. Structured retrieval has also been explored in enterprise workflows through graph-based indexing and semantic retrieval mechanisms~\citep{su2026grag4pm}. However, retrieval effectiveness alone does not establish whether the selected evidence is legally applicable or adequately supports individual claims.

\section{Trustworthiness Evaluation Framework}
\label{sec:framework}
\subsection{Evaluation Unit and Procedure}
We evaluate trustworthiness at the level of a complete system response rather than the generated answer alone. For a question $q_i$, the system produces
\begin{equation}
o_i = (a_i, e_i, m_i),
\end{equation}
where $a_i$ denotes the answer or explanation, $e_i$ denotes the supporting evidence, and $m_i$ denotes the response mode. We consider four possible response modes:
\begin{equation}
m_i \in \{\text{answer},\text{clarify},\text{abstain},\text{refer}\}.
\end{equation}

Trustworthiness is characterised through three complementary dimensions,
\begin{equation}
T_i =
\left(
C^{\mathrm{legal}}_i,
I^{\mathrm{evidence}}_i,
S^{\mathrm{decision}}_i
\right),
\end{equation}
where $C^{\mathrm{legal}}$ represents legal correctness, $I^{\mathrm{evidence}}$ represents evidence integrity, and $S^{\mathrm{decision}}$ represents decision safety. These dimensions are reported separately rather than combined into a weighted score. 

The evaluation requirements depend on the expected response mode. Let $m_i^{\star}$ denote the expected mode for question $q_i$. We define a requirement vector
\begin{equation}
\mathbf{r}_i =
g(m_i^{\star})
=
\left(
r_i^{\mathrm{legal}},
r_i^{\mathrm{evidence}},
r_i^{\mathrm{decision}}
\right),
\qquad
r_i^{(k)} \in \{0,1\},
\label{eq:requirement}
\end{equation}
where $r_i^{(k)}=1$ indicates that dimension $k$ is required for evaluating the response. The corresponding set of required dimensions is
\begin{equation}
\mathcal{R}_i =
\{k : r_i^{(k)} = 1\}.
\label{eq:requiredset}
\end{equation}

This formulation allows the evaluation to reflect the expected behaviour of each case. A direct answer requires assessment of its legal content, supporting evidence, and response decision. By contrast, an appropriate abstention should not be penalised for failing to provide a substantive legal conclusion, but it must still satisfy the criteria associated with the decision to abstain.

For each required dimension $k \in \mathcal{R}_i$, we define $z_i^{(k)} \in \{0,1\}$ to indicate whether the corresponding evaluation
criteria are satisfied. These indicators are determined from the applicable question level criteria and evaluation annotations described in Section~\ref{sec:experiments}, rather than from an arbitrary weighted combination of the three dimensions.

\begin{figure}[!htbp]
    \centering
    \includegraphics[width=0.9\linewidth]{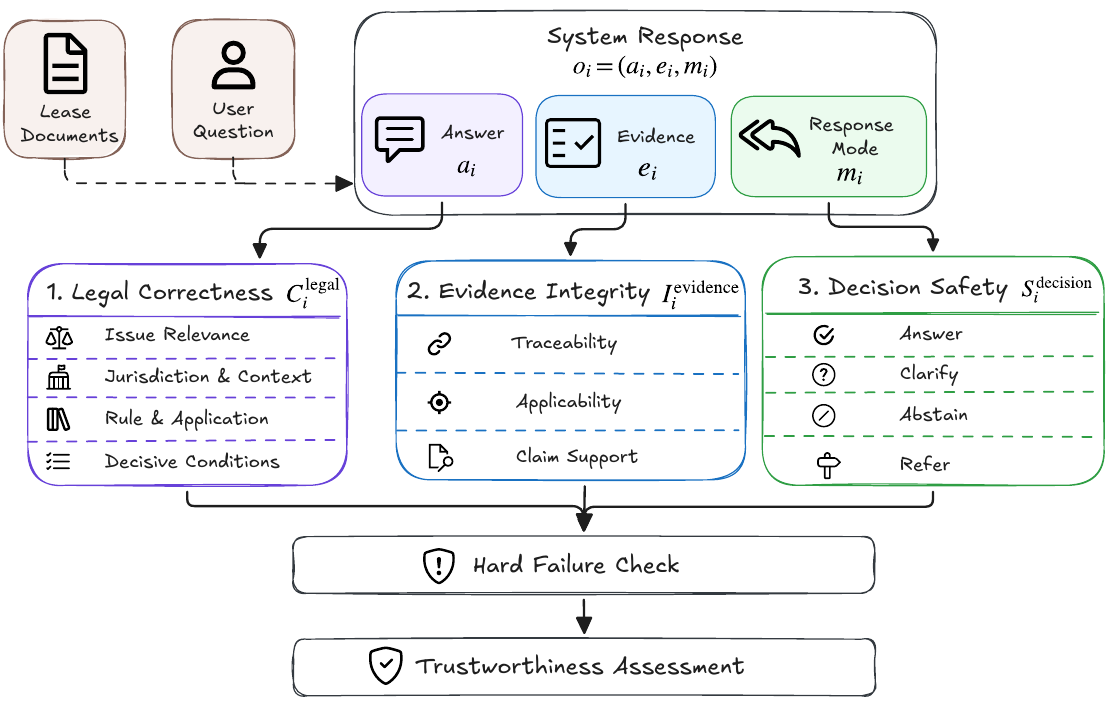}
    \caption{Trustworthiness evaluation framework.}
    \label{fig:trust_framework}
\end{figure}

The procedure is therefore hierarchical. We first identify the expected response mode and its required evaluation dimensions. We then assess the relevant dimensions and finally apply a separate hard failure gate. This distinguishes an incorrect answer from an incorrect decision to answer and prevents strong performance in one dimension from compensating for a serious failure in another. Figure~\ref{fig:trust_framework} summarises the evaluation procedure.

\subsection{Legal Correctness}

Legal correctness assesses whether the response identifies and applies the relevant legal context appropriately. We evaluate four aspects. \textbf{Issue relevance} assesses whether the response addresses the legal issue raised by the question rather than a related but different issue. \textbf{Jurisdiction and context} assess whether the response uses the appropriate jurisdiction and tenancy setting. \textbf{Rule and application} assess whether substantive or procedural statements are consistent with the applicable legal material. \textbf{Decisive conditions} assess whether the response recognises conditions, exceptions, or missing facts that could materially change the answer.

A response may reach the expected conclusion for the wrong reason. For example, it may rely on material from an inapplicable tenancy context or omit a condition that determines whether the stated rule applies. Such a response should not receive the same evaluation
as one that reaches the conclusion using the appropriate legal context.

\subsection{Evidence Integrity}

Evidence integrity assesses whether the supporting material is suitable for the claims made in the response. We distinguish three properties. \textbf{Traceability} asks whether cited material can be located in the underlying source collection. \textbf{Applicability} asks whether the material is appropriate for the relevant jurisdiction, tenancy type, and legal issue. \textbf{Support} asks whether the cited material actually supports the associated claim. A citation may therefore be traceable but inapplicable, or applicable but insufficient to support the claim attributed to it. The presence of a citation alone is consequently not treated as evidence integrity.

For evaluation, we define a \emph{material claim} as a statement concerning a legal rule, obligation, entitlement, condition, deadline, amount, remedy, procedural step, or recommended action whose correctness could affect the user's interpretation or next step. Each material claim is assessed according to whether it is supported, partially supported, or unsupported by the cited evidence. This claim level representation provides the basis for the Evidence Support measure used in Section~\ref{sec:experiments}.

Evidence Support is computed at the material claim level as the proportion of material claims classified as supported by their cited evidence. Partially supported and unsupported claims are not counted as fully supported.

\subsection{Decision Safety}

Decision safety evaluates whether the system selects an appropriate response mode. In high stakes legal information tasks, providing a direct answer is not always the correct behaviour. An \textbf{answer} is appropriate when sufficient applicable information is available to address the question. \textbf{Clarify} is appropriate when a decisive fact is missing and answering would otherwise require an unsupported assumption. \textbf{Abstain} is appropriate when the request falls outside the supported scope or cannot be answered reliably from the available information. \textbf{Refer} is appropriate when the circumstances warrant further assistance rather than reliance on an automated explanation alone.

This dimension therefore evaluates the decision to answer separately from the quality of the generated text. A fluent and factually plausible response can still be unsafe if the correct behaviour was to request clarification, abstain, or refer.

\subsection{Hard Failure Gate}

Some errors are sufficiently serious that averaging performance across dimensions would obscure their practical importance. We therefore apply a separate hard failure gate after the dimension level assessment.

We consider $R$ predefined categories of serious failure. Let $f_{ir} \in \{0,1\}$ indicate whether response $i$ contains failure type $r$. The response level hard failure indicator is
\begin{equation}
h_i =
\mathbb{I}
\left(
\sum_{r=1}^{R} f_{ir} > 0
\right),
\label{eq:hardfailure}
\end{equation}
so that any single serious failure is sufficient to activate the gate.

The current evaluation considers the following categories: 1) materially incorrect legal information; 2) use of the wrong jurisdiction or tenancy context; 3) a fabricated or untraceable citation; 4) an unsupported deadline, amount, remedy, power, or obligation; 5) a dangerous omission or unsafe action; 6) unsupported certainty about a material legal issue; and 7) failure to clarify, abstain, or refer when such behaviour is required.

We then define a response level trustworthiness indicator as
\begin{equation}
\tau_i =
(1-h_i)
\prod_{k \in \mathcal{R}_i}
z_i^{(k)},
\qquad
\tau_i \in \{0,1\}.
\label{eq:trustworthy}
\end{equation}

A response therefore satisfies the evaluation protocol only when no hard failure is present and all dimensions required for that case satisfy their evaluation criteria. This definition is deliberately non compensatory: strong performance in one dimension cannot offset failure in another required dimension or the presence of a hard failure.

At the dataset level, we report both the hard failure rate and the trustworthy
response rate:
\begin{equation}
\mathrm{HFR}
=
\frac{1}{N}\sum_{i=1}^{N} h_i,
\qquad
\mathrm{TRR}
=
\frac{1}{N}\sum_{i=1}^{N} \tau_i.
\label{eq:aggregate}
\end{equation}

HFR measures how frequently serious failures occur, whereas TRR measures the proportion of responses that satisfy all required trustworthiness criteria without triggering the hard failure gate. These measures complement conventional answer accuracy by distinguishing an apparently correct answer from a response that is suitable under the complete evaluation protocol.

\section{Experiments and Results}
\label{sec:experiments}

\subsection{Experimental Setting}

We evaluate the framework on a benchmark of 2,000 New South Wales residential tenancy questions spanning five broad tenancy topics, multiple question types, risk levels, and expected response modes. We use a family preserving split to reduce leakage between closely related questions derived from the same legal issue. The resulting partition contains 1,400 training, 300 validation, and
300 protected test questions. All development decisions, including prompt design, retrieval settings, threshold selection, and response rules, are made using only the training and validation sets. The model configuration, legal source collection, prompts, and evaluation procedure are frozen before the protected test set is examined. Additional benchmark and provenance statistics are provided in the appendix.

All configurations use the same Llama 3.1 8B Instruct model~\citep{dubey2024llama3} with identical inference settings and experimental conditions. We compare five progressively safeguarded configurations. \textbf{Direct Generation} generates an answer from the base retrieved evidence without applicability or post-generation checks. \textbf{Applicability Aware Retrieval} adds checks for jurisdiction, tenancy context, and legal topic before evidence selection. \textbf{Evidence First Generation} constructs a conservative response from selected evidence before optional language generation. \textbf{Evidence Checked Generation} further checks material statements against cited evidence. The \textbf{Full Safeguarded System} adds fallback behaviour and explicit
selection among answer, clarify, abstain, and refer.

\paragraph{Evaluation procedure.}
Two legal experts independently annotate protected test responses against the predefined criteria for legal correctness, evidence support, response mode, and hard failures. Disagreements are reviewed before the final labels are fixed, and automated checks support source traceability and evidence assessment. Mean inter annotator agreement across categorical evaluation labels is substantial (Cohen's $\kappa=0.78$).

We report Answer Accuracy as the conventional performance measure. Answer Accuracy is computed over all 300 protected test questions as the proportion for which the substantive answer outcome matches the predefined reference outcome, independently of the expected response mode. The response level measures are Trustworthy Response Rate (TRR), Hard Failure Rate (HFR), Evidence Support, and Response Mode Accuracy. TRR, HFR, Answer Accuracy, and Response Mode Accuracy are computed over the 300 question protected test set, while
Evidence Support is the proportion of material claims classified as supported by their cited evidence. For binary response level measures, we report 95\% Wilson score confidence intervals. Since all configurations are evaluated on the same questions, paired differences are assessed using exact McNemar tests.

\subsection{Accuracy versus Trustworthiness}
We examine whether conclusions based on Answer Accuracy remain stable when the complete response is evaluated. Table~\ref{tab:main_results} reports the results across the five system configurations.

\begin{table*}[!htbp]
\centering
\caption{Performance on the 300 question protected test set under conventional and trustworthiness measures.}
\label{tab:main_results}
\small
\setlength{\tabcolsep}{7pt}
\begin{tabular}{lccccc}
\toprule
System &
Accuracy $\uparrow$ &
TRR $\uparrow$ &
HFR $\downarrow$ &
Evidence Support $\uparrow$ &
Mode Acc. $\uparrow$ \\
\midrule
Direct Generation   & 82.3 & 58.7 & 20.7 & 70.2 & 68.3 \\
Applicability Aware & 83.0 & 65.7 & 15.7 & 77.8 & 68.7 \\
Evidence First      & 83.7 & 72.0 & 11.7 & 84.5 & 69.0 \\
Evidence Checked    & 83.7 & 78.0 & 8.0  & 91.2 & 69.0 \\
\textbf{Full Safeguarded}    & \textbf{84.0} & \textbf{82.3} & \textbf{5.7}  & \textbf{92.0} & \textbf{88.7} \\
\bottomrule
\end{tabular}
\end{table*}

From Direct Generation to the Full Safeguarded System, Answer Accuracy increases from 82.3\% to 84.0\%, whereas TRR increases from 58.7\% to 82.3\% and HFR decreases from 20.7\% to 5.7\%. Evidence Support increases from 70.2\% to 92.0\%, a difference of 21.8 percentage points. The uncertainty estimates show a similar pattern. Answer Accuracy is 82.3\% (95\% CI: 77.6--86.2) for Direct Generation and 84.0\%
(79.4--87.7) for the Full Safeguarded System. The paired difference in Answer Accuracy is not statistically significant ($p>0.05$). In contrast, TRR increases from 58.7\% (53.0--64.1) to 82.3\% (77.6--86.2), with a significant paired difference ($p<10^{-7}$), while HFR decreases from 20.7\% (16.5--25.6) to 5.7\% (3.6--8.9) ($p<10^{-6}$). Response Mode Accuracy also increases from 68.3\%
(62.9--73.3) to 88.7\% (84.6--91.8) ($p<10^{-7}$).

The difference is also visible between Evidence First Generation and Evidence Checked Generation. Both configurations achieve 83.7\% Answer Accuracy, while TRR increases from 72.0\% to 78.0\%, HFR decreases from 11.7\% to 8.0\%, and Evidence Support increases from 84.5\% to 91.2\%. Thus, configurations that are indistinguishable under Answer Accuracy can still differ in evidence support and response level trustworthiness.

The system ordering remains broadly consistent under Answer Accuracy and TRR, with the Full Safeguarded System performing best under both measures. However, the differences between configurations are substantially larger under TRR and HFR than under Answer Accuracy.

\subsection{Effects of Safeguards}

We examine the changes associated with each successive safeguard. Adding \textbf{Applicability Aware Retrieval} increases TRR from 58.7\% to 65.7\%, reduces HFR from 20.7\% to 15.7\%, and increases Evidence Support from 70.2\% to 77.8\%, while Answer Accuracy changes only from 82.3\% to 83.0\%.

Adding \textbf{Evidence First Generation} further increases TRR to 72.0\% and Evidence Support to 84.5\%, while HFR decreases to 11.7\%. Answer Accuracy increases by only 0.7 percentage points to 83.7\%. \textbf{Evidence Checked Generation} leaves Answer Accuracy unchanged at 83.7\%, but increases TRR from 72.0\% to 78.0\% and Evidence Support from 84.5\% to 91.2\%, while HFR decreases from 11.7\% to 8.0\%. Configurations with identical Answer Accuracy can therefore differ substantially under response level trustworthiness measures.

The \textbf{Full Safeguarded System} increases Response Mode Accuracy from 69.0\% to 88.7\%, while Answer Accuracy changes only from 83.7\% to 84.0\%. Successive safeguards are associated with substantially larger changes in the response level measures than in Answer Accuracy.

\subsection{Slice Analysis}

\begin{figure}[!htbp]
    \centering

    \begin{subfigure}[t]{0.49\textwidth}
        \centering
        \includegraphics[width=\linewidth]{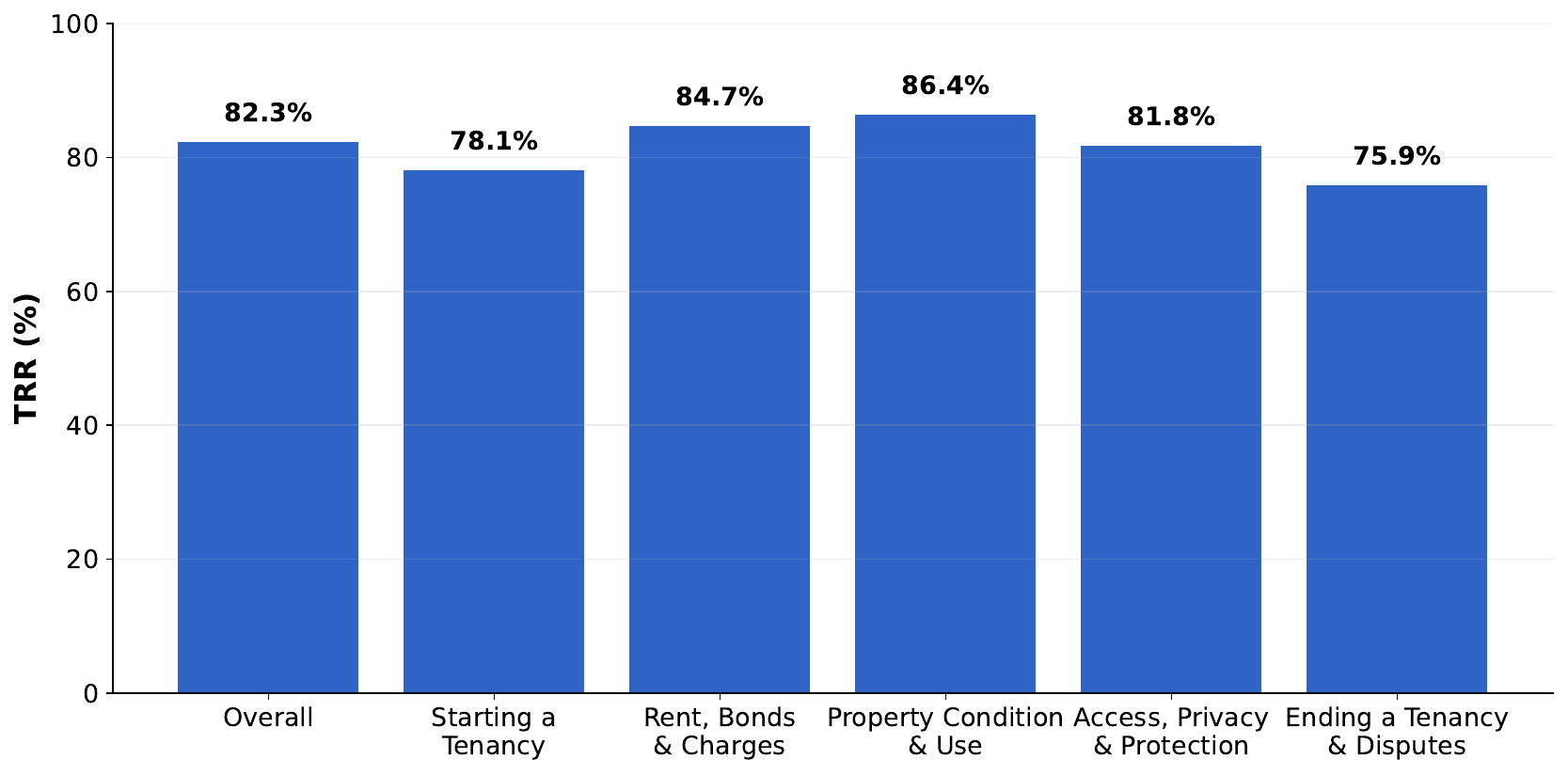}
        \caption{Trustworthy Response Rate across topics.}
        \label{fig:trr_topics}
    \end{subfigure}
    \hfill
    \begin{subfigure}[t]{0.49\textwidth}
        \centering
        \includegraphics[width=\linewidth]{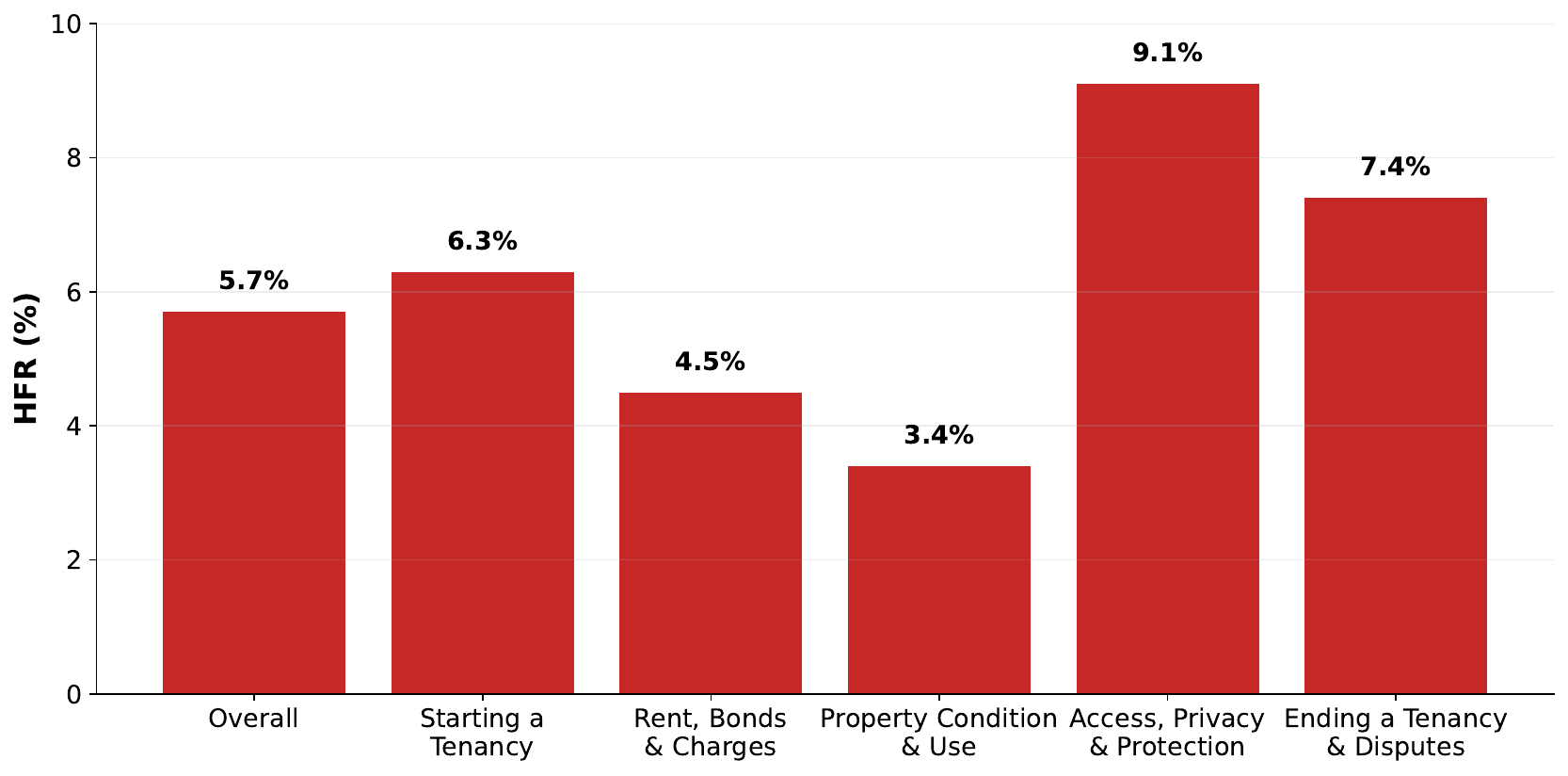}
        \caption{Hard Failure Rate across topics.}
        \label{fig:hfr_topics}
    \end{subfigure}

    \vspace{0.5em}

    \begin{subfigure}[t]{0.65\textwidth}
        \centering
        \includegraphics[width=\linewidth]{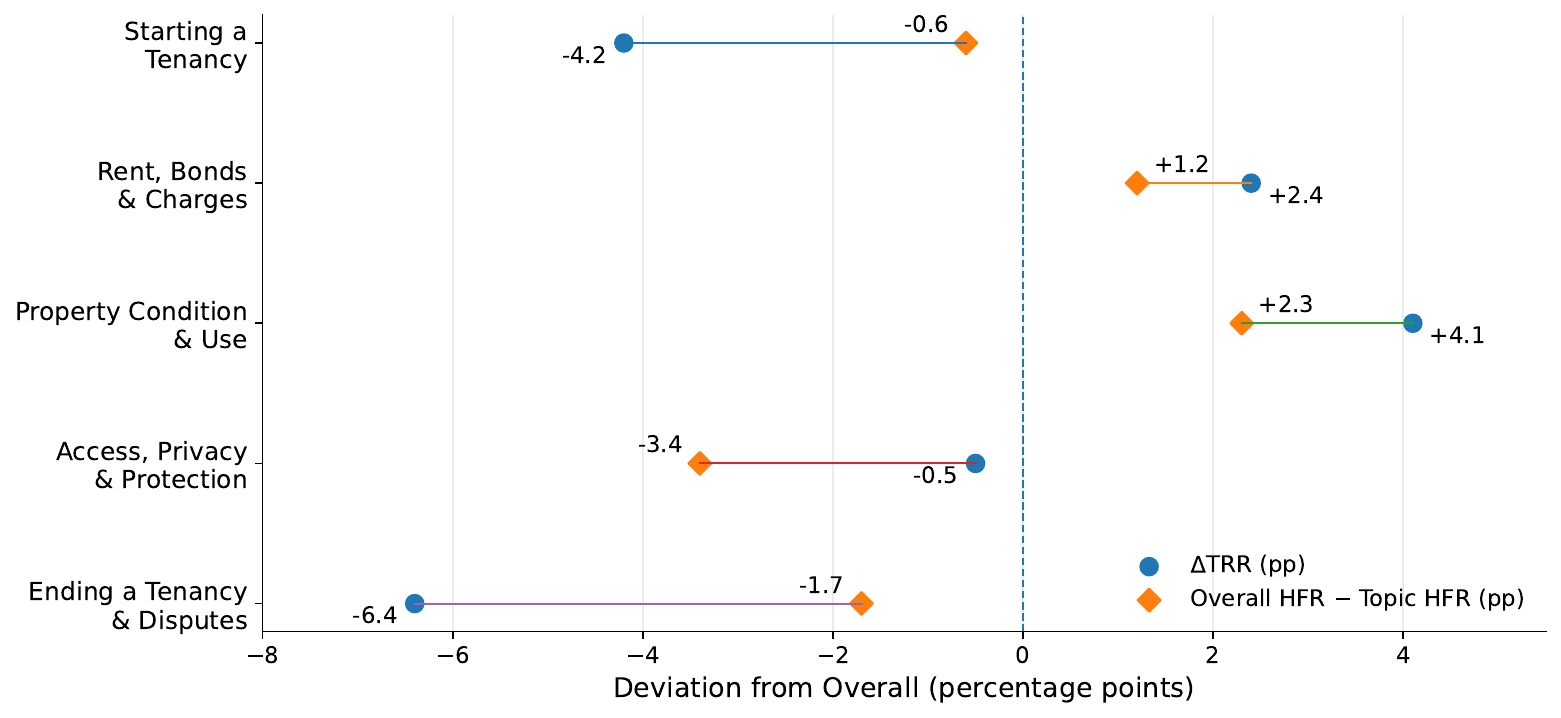}
        \caption{Deviation from overall performance across tenancy topics.}
        \label{fig:topic_gap}
    \end{subfigure}

    \caption{Topic level slice analysis of the Full Safeguarded System. In (c), positive values indicate better than overall performance for both measures.}
    \label{fig:worst_slice}
\end{figure}

We examine the Full Safeguarded System across the five tenancy topics to assess whether aggregate performance is consistent across different areas of the benchmark. Figure~\ref{fig:worst_slice} reports topic level TRR and HFR, together with their deviations from the overall results.

The Full Safeguarded System achieves an overall TRR of 82.3\%. Topic level TRR ranges from 75.9\% to 86.4\%. \textbf{Ending a Tenancy \& Disputes} records the lowest TRR at 75.9\%, 6.4 percentage points below the overall result, followed by \textbf{Starting a Tenancy} at 78.1\%. In contrast, \textbf{Property Condition \& Use} achieves the highest TRR at 86.4\%, 4.1 points above the overall result.

The overall HFR is 5.7\%, with topic level values ranging from 3.4\% to 9.1\%. \textbf{Access, Privacy \& Protection} records the highest HFR at 9.1\%, 3.4 percentage points worse than the overall result, while \textbf{Property Condition \& Use} records the lowest HFR at 3.4\%. \textbf{Rent, Bonds \& Charges} also has a higher TRR and lower HFR than the aggregate result.

Figure~\ref{fig:worst_slice}(c) highlights that the two measures reveal different weakness profiles. \textbf{Ending a Tenancy \& Disputes} shows the largest negative deviation in TRR, whereas \textbf{Access, Privacy \& Protection} shows the largest deterioration in HFR despite a TRR close to the overall result. By contrast, \textbf{Property Condition \& Use} performs better than the overall system under both measures. These differences show that aggregate performance alone can obscure topic specific patterns of response level trustworthiness.

\section{Discussion, Limitations, and Conclusion}
Answer Accuracy and response level trustworthiness can yield different assessments of otherwise similar system configurations. A
particularly clear example is the comparison between Evidence First Generation and Evidence Checked Generation: both achieve 83.7\% Answer Accuracy, yet their TRR and HFR differ substantially. This suggests that conventional accuracy can overlook differences in whether responses satisfy evidence and safety requirements. The topic analysis reveals a complementary pattern: the topic with the lowest TRR is not the same as the topic with the highest HFR, indicating that these measures expose different weakness profiles that are
not visible from aggregate performance alone.

Several limitations should be considered. The evaluation is restricted to New South Wales residential tenancy and does not establish reliability across other jurisdictions or legal domains. The legal source collection is frozen at a specific date, so subsequent legislative or policy changes are outside the evaluation scope. Although the 2,000 question benchmark spans five broad tenancy topics, multiple question types, risk levels, and response requirements, structured benchmark questions remain simpler than many real renter narratives. Some aspects of legal correctness and evidence support require expert judgement, while automated evidence checks provide conservative approximations rather than complete legal reasoning. The three evaluation dimensions provide an operational definition of trustworthiness for this setting rather than a universal definition of trustworthy legal AI. Even a safeguarded system may create risks of overreliance or delayed professional assistance; its outputs should therefore be treated as legal information rather than a substitute for legal advice. Finally, the study evaluates system responses rather than downstream user decisions or real world legal outcomes.

Within these limits, the results support evaluating high stakes lease interpretation at the level of the complete response rather than answer accuracy alone. Evaluation should also consider whether the legal conclusion is correct, whether supporting evidence is applicable and sufficient, whether serious failures occur, and whether the system selects a suitable response behaviour.



{\small
\bibliographystyle{plainnat}
\bibliography{ref}
}


\appendix

\section{Additional Experimental Details}

\subsection{Benchmark Composition}

The benchmark contains 2,000 New South Wales residential tenancy questions across five broad topics: Starting a Tenancy (215), Rent, Bonds \& Charges (740), Property Condition \& Use (395), Access, Privacy \& Protection (290), and Ending a Tenancy \& Disputes (360). Questions derived from the same underlying legal issue are grouped into families and assigned to the same partition. The resulting split contains 1,400 training, 300 validation, and 300 protected test questions.

The legal source collection is restricted to New South Wales residential tenancy materials and is fixed before protected test evaluation. The test set is not used for prompt design, retrieval tuning, threshold selection, or response rule development.

\subsection{Evaluation Details}

Two legal experts independently assess protected test responses using the criteria defined in Section~\ref{sec:framework}. Disagreements are reviewed before the final labels are fixed. Inter annotator agreement across categorical evaluation labels is substantial, with Cohen's $\kappa=0.78$.

Material claims are labelled as supported, partially supported, or unsupported. Evidence Support is the proportion of material claims classified as fully supported. Answer Accuracy is computed over the 300 protected test questions using the predefined substantive reference outcome, independently of evidence quality and response mode.

\subsection{Model and Implementation Details}

All configurations use Meta Llama 3.1 8B Instruct
(\texttt{meta-llama/Llama-3.1-8B-Instruct}) \citep{dubey2024llama3}, an instruction tuned, text only 8B model with a 128K context window. The same model and inference settings are used across all configurations. The model is distributed under the Llama 3.1 Community License.

\subsection{Statistical Results}

For binary response level measures, 95\% confidence intervals are calculated using Wilson score intervals. Paired differences between configurations are assessed using exact McNemar tests because all configurations are evaluated on the same protected test questions.

\begin{table}[h]
\centering
\caption{Performance with 95\% Wilson score confidence intervals on the
protected test set. Values are percentages.}
\label{tab:appendix_ci}
\small
\setlength{\tabcolsep}{4pt}
\begin{tabular}{lcccc}
\toprule
System & Accuracy & TRR & HFR & Mode Acc. \\
\midrule
Direct Generation
& 82.3 [77.6, 86.2]
& 58.7 [53.0, 64.1]
& 20.7 [16.5, 25.6]
& 68.3 [62.9, 73.3] \\

Applicability Aware
& 83.0 [78.3, 86.8]
& 65.7 [60.1, 70.8]
& 15.7 [12.0, 20.2]
& 68.7 [63.2, 73.7] \\

Evidence First
& 83.7 [79.1, 87.4]
& 72.0 [66.7, 76.8]
& 11.7 [8.5, 15.8]
& 69.0 [63.6, 74.0] \\

Evidence Checked
& 83.7 [79.1, 87.4]
& 78.0 [73.0, 82.3]
& 8.0 [5.4, 11.6]
& 69.0 [63.6, 74.0] \\

Full Safeguarded
& 84.0 [79.4, 87.7]
& 82.3 [77.6, 86.2]
& 5.7 [3.6, 8.9]
& 88.7 [84.6, 91.8] \\
\bottomrule
\end{tabular}
\end{table}

For Direct Generation versus the Full Safeguarded System, the paired change in Answer Accuracy is not statistically significant ($p>0.05$), whereas the changes in TRR ($p<10^{-7}$), HFR ($p<10^{-6}$), and Response Mode Accuracy ($p<10^{-7}$) are statistically significant.



\end{document}